\documentclass[final]{cvpr}

\usepackage{times}
\usepackage{epsfig}
\usepackage{graphicx}
\usepackage{amsmath}
\usepackage{amssymb}
\usepackage{comment}
\newcommand{\bfsection}[1]{\vspace*{0.1cm}\noindent\textbf{#1.}}
\usepackage{multirow}
\usepackage[pagebackref=true,breaklinks=true,colorlinks,bookmarks=false]{hyperref}

\def\cvprPaperID{9218} 
\def\confYear{CVPR 2022}

\usepackage{helvet} 
\usepackage{courier}  
\usepackage[square,numbers]{natbib}
\usepackage[utf8]{inputenc} 
\usepackage[T1]{fontenc}    
\usepackage{hyperref}
\usepackage{url}

\usepackage{booktabs}       
\usepackage{amsfonts}       
\usepackage{nicefrac}       
\usepackage{microtype}      

\usepackage{epsfig}
\usepackage{amsmath}
\usepackage{amssymb}
\usepackage{adjustbox}

\usepackage{amsthm}  
\usepackage{wasysym}
\usepackage[ruled,vlined,algo2e]{algorithm2e}

\usepackage{color}
\usepackage{dsfont}	
\usepackage{wrapfig}
\usepackage{footnote}
\usepackage{epstopdf}
 \usepackage{enumitem}
\usepackage{caption}
\usepackage{comment}
\usepackage{multirow}
\usepackage{algorithm}
\usepackage{algorithmic}
\usepackage{mathtools}
\usepackage{makecell}
\usepackage{subfig}
\usepackage{balance}

\def\eg{\emph{e.g., }}
\def\ie{\emph{i.e., }}

\def\etal{\emph{et al. }}

\def\stackrel#1#2{\mathrel{\mathop{#2}\limits^{#1}}}

\usepackage{pifont}

\DeclareMathOperator*{\argmin}{arg\,min}

\usepackage{mathtools}

\makeatletter
\newcommand*{\rom}[1]{\expandafter\@slowromancap\romannumeral #1@}
\makeatother
\makeatletter
\newcommand\footnoteref[1]{\protected@xdef\@thefnmark{\ref{#1}}\@footnotemark}
\makeatother

\newtheorem{prop}{Proposition}

\begin{document}

\title{Robust Structured Declarative Classifiers for 3D Point Clouds: Defending Adversarial Attacks with Implicit Gradients}




\author{Kaidong Li\textsuperscript{\rm 1}, \quad Ziming Zhang\textsuperscript{\rm 2}\thanks{Joint first author}, \quad Cuncong Zhong\textsuperscript{\rm 1}, \quad Guanghui Wang\textsuperscript{\rm 3}\\
\textsuperscript{\rm 1}Department of EECS, University of Kansas, KS, USA \\ 
\textsuperscript{\rm 2}Department of ECE, Worcester Polytechnic Institute, MA, USA\\
\textsuperscript{\rm 3}Department of CS, Ryerson University, Toronto ON, Canada \\
{\tt\small \{kaidong.li, cczhong\}@ku.edu, zzhang15@wpi.edu, wangcs@ryerson.ca}
}

\maketitle
\pagestyle{empty}
\thispagestyle{empty}

\begin{abstract}
Deep neural networks for 3D point cloud classification, such as PointNet, have been demonstrated to be vulnerable to adversarial attacks. Current adversarial defenders often learn to denoise the (attacked) point clouds by reconstruction, and then feed them to the classifiers as input. In contrast to the literature, we propose a family of robust structured declarative classifiers for point cloud classification, where the internal constrained optimization mechanism can effectively defend adversarial attacks through implicit gradients. Such classifiers can be formulated using a bilevel optimization framework. We further propose an effective and efficient instantiation of our approach, namely, {\em Lattice Point Classifier (LPC)}, based on structured sparse coding in the permutohedral lattice and 2D convolutional neural networks (CNNs) that is end-to-end trainable. We demonstrate state-of-the-art robust point cloud classification performance on ModelNet40 and ScanNet under seven different attackers. For instance, we achieve {\bf 89.51\%} and {\bf 83.16\%} test accuracy on each dataset under the recent JGBA attacker that outperforms DUP-Net and IF-Defense with PointNet by $\sim$70\%. The demo code is available at \url{https://zhang-vislab.github.io}.

\end{abstract}

\section{Introduction}
Point clouds are unstructured data which is widely used in real-world applications such as autonomous driving. To recognize them using deep neural networks, point clouds can be represented as points \cite{qi2017pointnet}, images \cite{lyu2020learning}, voxels \cite{wu20153d}, or graphs \cite{wang2019dynamic}. Recent works \cite{zhou2020lg, Xiang_2021_ICCV, ma2020efficient, guo2019simple, xiang2019generating, wicker2019robustness, liu2020adversarial} have demonstrated that such deep networks are vulnerable to (gradient-based) adversarial attacks. 
Accordingly, several adversarial defenders \cite{zhou2019dup, wu2020if, liu2021pointguard} have been proposed for robust point cloud classification. The basic ideas are often to denoise the (attacked) point clouds before feeding them into the classifiers as input to preserve their prediction accuracy. 

\begin{figure}[t]
    \includegraphics[width=1\linewidth]{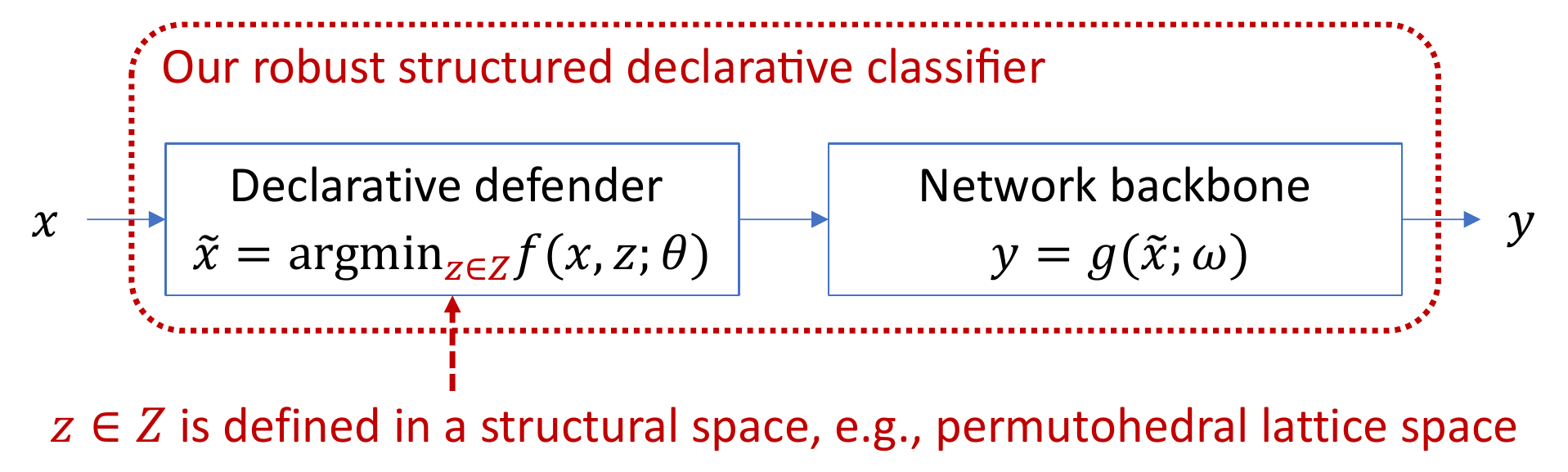}
    \vspace{-7mm}
    \caption{Illustration of robust structured declarative classifiers, where our defender is optimized in a structural space.} 
    \label{fig:framework}
    \vspace{-4mm}
\end{figure}

\bfsection{Obfuscated gradients}
In the white-box adversarial attacks, the attackers are assumed to have full access to both classifiers and defenders. To defend such attacks, one common way is to {\em break} the gradient over the input data in the backpropagation (either inadvertently or intentionally, \eg a defender is non-differentiable or prevents gradient signal from flowing through the network) so that the attackers fail to be optimized. Such scenarios are called obfuscated gradients. In \cite{athalye2018obfuscated} Athalye \etal have discussed the false sense of security in such defenders and proposed new methods, such as Backward Pass Differentiable Approximation (BPDA), to attack them successfully. Take DUP-Net \cite{zhou2019dup} for example, where a non-differentiable Statistical Outlier Removal (SOR) defense strategy was proposed. In \cite{ma2020efficient} Ma \etal proposed Joint Gradient Based Attack (JGBA) that can compute the gradient with a linear approximation (an instantiation of BPDA) of the SOR defense to attack DUP-Net successfully.

\bfsection{Implicit gradients}
Now let us consider the scenarios where both defenders and classifiers are differentiable. Then in order to defend the adversarial attacks, one way is to make the calculation of the gradient challenging. To this end, implicit gradients \cite{rajeswaran2019meta} may be more suitable for designing the defenders. {\em An implicit gradient, $\frac{\partial y}{\partial x}$, is defined by a differentiable function $h$ that takes $x, y$ as its input, \ie $\frac{\partial y}{\partial x} = h(x, y)$.} Such an equation can be also considered as a first-order ordinary differential equation (ODE), which is solvable (approximately) using Euler's Method \cite{kag2019rnns}. Here we assume that the gradients through the classifiers 
can be easily computed, which often holds empirically. To our best knowledge, so far there is no work on designing adversarial defenders for 3D point clouds based on implicit gradients.

\bfsection{Declarative networks}
Implicit gradients require equations that contain both the input and output of a defender. One potential solution for this is to introduce optimization problems as the defenders, where the first-order optimality conditions provide such equations. In the literature, there have been some works \cite{amos2017optnet, agrawal2019differentiable, lee2019meta, rajeswaran2019meta} that proposed optimization as network layers in deep neural networks. Recently in \cite{gould2021deep} Gould \etal generalized these ideas and proposed deep declarative networks. A {\em declarative} network node is introduced where the exact implementation of the forward processing function is not defined; rather the input-output relationship $(x\mapsto \Tilde{x})$ is defined in terms of behavior specified as the solution to an optimization problem $\Tilde{x}\in\argmin_{z\in \mathcal{Z}} f(x,z;\theta)$. Here $f$ is an objective function, $\theta$ denotes the node parameters, and $\mathcal{Z}$ is the feasible solution space. In \cite{gould2021deep} a robust pooling layer was proposed as a declarative node using unconstrained minimization with various penalty functions such as Huber or Welsch, which can be efficiently solved using Newton's method or gradient descent. The effectiveness of such pooling layers was demonstrated for point cloud classification.

\begin{figure}[t]
    \centering
    \includegraphics[width=0.9\linewidth]{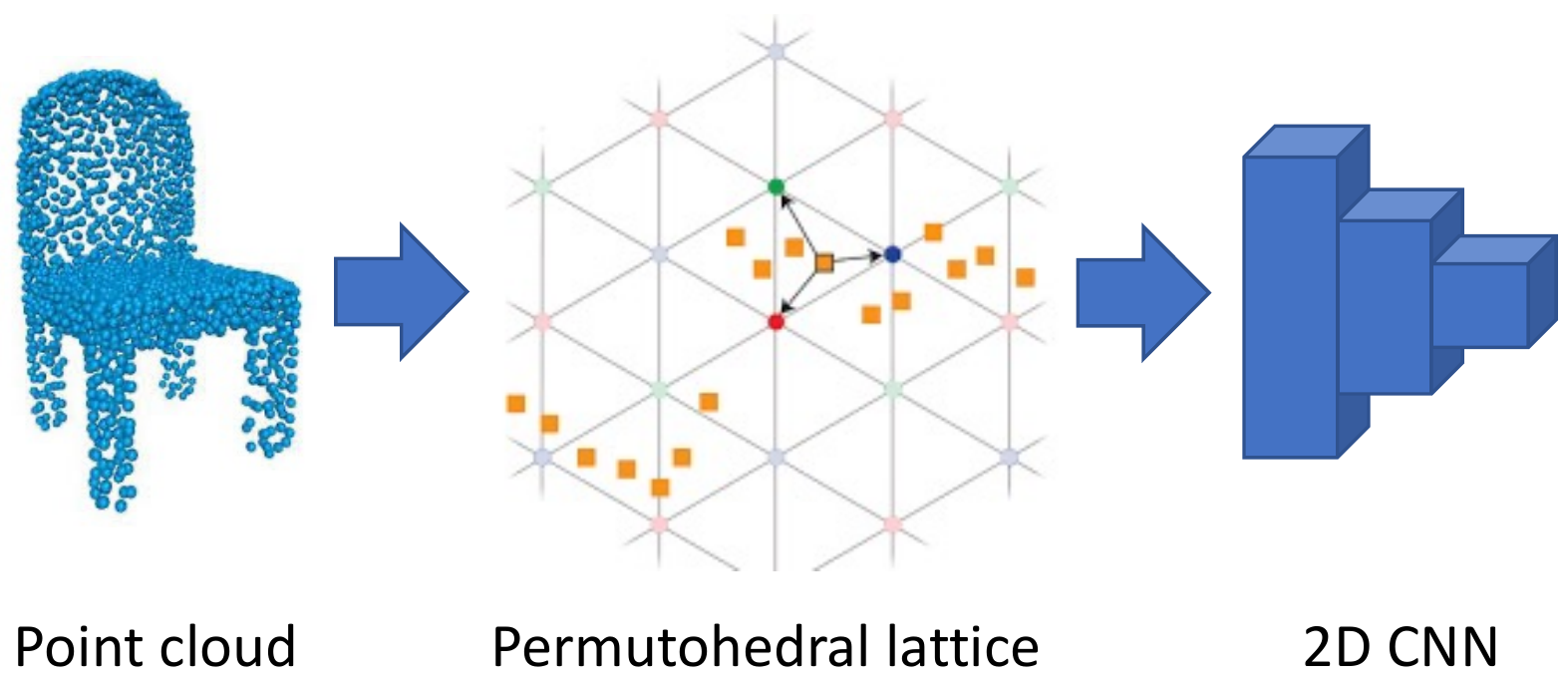}
    \vspace{-2mm}
    \caption{Illustration of our Lattice Point Classifier (LPC). 
    } 
    \label{fig:LPC}
    \vspace{-4mm}
\end{figure}

\bfsection{Our approach}
Motivated by the methods above, in this paper we propose a novel robust structured declarative classifiers for 3D point clouds by embedding a declarative node into the networks, as illustrated by Fig. \ref{fig:framework}. Different from robust pooling layers in \cite{gould2021deep}, our declarative defender is designed to reconstruct each point cloud in a (learnable) structural space as a means of denoising. To this end, we borrow the idea from structured sparse coding \cite{szlam2012fast, karygianni2014structured, wei2017structured} by representing each point as a linear combination of atoms in a dictionary. Together with the backbone networks, the training of our robust classifiers can lead to a bilevel optimization problem. Considering the inference efficiency, one plausible instantiation of our classifiers, as illustrated in Fig.~\ref{fig:LPC}, is to define the structural space using the permutohedral lattice \cite{adams2010fast, su2018splatnet, gu2019hplflownet}, project each point cloud onto the lattice, generate a 2D image based on the barycentric weights, and feed the image to a 2D convolutional neural network (CNN) for classification. We call this instantiation {\em Lattice Point Classifier (LPC)}. 

\bfsection{Our contributions}
We summarize our contributions below:
\begin{itemize}[nosep, leftmargin=*]
    \item We propose a family of novel robust structured declarative classifiers for 3D point clouds where the declarative nodes defend the adversarial attacks through implicit gradients. To the best of our knowledge, we are the {\em first} to explore implicit gradients in robust point cloud classification.
    
    \item We propose a bilevel optimization framework to learn the network parameters in an end-to-end fashion.
    
    \item We propose an effective and efficient instantiation of our robust classifiers based on the structured sparse coding in the permutohedral lattice and 2D CNNs.
    
    \item We demonstrate superior performance of our approach by comparing with the state-of-the-art adversarial defenders under the state-of-the-art adversarial attackers.
\end{itemize}

	

\section{Related Works}\label{sect:related}
\bfsection{Deep learning for 3D point clouds}
Based on the point cloud representations, we simply group some typical deep networks into four categories. \textit{Point-based networks} \cite{qi2017pointnet, qi2017pointnet++, li2018pointcnn, yan2020pointasnl, wu2019unsupervised} directly take each point cloud as input, extract point-wise features using multi-layer perceptrons (MLPs), and fuse them to generate a feature for the point cloud. \textit{Image-based networks} \cite{su2015multi, qi2016volumetric, yu2018multi, yang2019learning, lyu2020learning, mo2021stereo} often project a 3D point cloud onto a (or multiple) 2D plane to generate a (or multiple) 2D image for further process. \textit{Voxel-based networks} \cite{maturana2015voxnet, wu20153d, riegler2017octnet, wang2017cnn, le2018pointgrid} usually voxelize each point cloud into a volumetric occupancy grid and further some classification techniques such as 3D CNNs are used for the tasks. {\em Graph-based networks} \cite{landrieu2019point, wang2019dynamic, shi2020point, fu2021robust, qian2021pu} often represent each point cloud as a graph such as KNN or adjacency graph which are fed to train graph convolutional networks (GCNs). A nice survey can be found in \cite{guo2020deep}.

\bfsection{Adversarial attacks on point clouds}
Such attacks aim to modify the input point clouds in a way that is not noticeable but can fool a classifier. Attackers can either have a target or not. Targeted attackers try to fool the classifier to predict a specified wrong class, while untargeted ones do not care about the predicted class as long as it is wrong. Nice surveys on adversarial attacks can be found in \cite{ozdag2018adversarial, wang2021adversarial, chakraborty2021survey}. Below we summarize some typical attackers for point clouds:
\begin{itemize}[nosep, leftmargin=*]
    \item {\em Point perturbation \cite{liu2019extending, yang2019adversarial, xiang2019generating, kim2021minimal, ma2020efficient}.} Inspired by Fast Gradient Sign Method (FGSM) \cite{goodfellow2014explaining}, they add on each point a small perturbation constrained by distance metrics (\eg $L_p$ norm \cite{xiang2019generating}, on the surface of an $\epsilon$-ball \cite{liu2019extending}).
    
    
    
    \item {\em Point addition \cite{xiang2019generating, wicker2019robustness, yang2019adversarial}.} Independent point attackers initially pick some points from target classes, add small perturbations and finally append these points to the victim point clouds. Cluster attackers similarly pick most critical points and then find a specified number of small point clusters to append to the victim point clouds. Object attackers attach scaled and rotated foreign object to the original point clouds. 
    
    \item {\em Point dropping.} Adversarial attackers pick critical points from each input point cloud, and drop them to fool the classifier. However, it is a non-differentiable operation. To address the issue in learning, \cite{zheng2019pointcloud} proposed creating saliency maps by viewing dropping a point as moving it to the cloud center. Wicker \etal \cite{wicker2019robustness} designed an algorithm to iteratively find the points to drop by minimizing a predefined objective function, similarly in \cite{yang2019adversarial}.
    
    \item {\em Others.} Hamdi \etal \cite{hamdi2020advpc} proposed a transferable adversarial perturbation attacker based on an adversarial loss that can learn the data distribution. Zhao \etal \cite{zhao2020isometry} proposed a black-box (\ie no access to the models) attacker with zero loss in isometry, as well as a white-box (\ie full access to the models) attacker based on the spectral norm. LG-GAN \cite{zhou2020lg} is generative adversarial network (GAN) based, which learns and incorporates target features into victim point clouds. A backdoor attacker was proposed in \cite{Xiang_2021_ICCV} to trick the 3D models by inserting adversarial point patterns into the training set so that the victim models learn to recognize the adversarial patterns during inference.  
\end{itemize}

\bfsection{Adversarial defense on point clouds} 
Adversarial defenses aim to denoise the input point clouds to recover the ground-truth labels from the classifiers. Nice surveys on adversarial defenses can be found in \cite{ozdag2018adversarial, wang2021adversarial, chakraborty2021survey}. Below we summarize some typical defenders for point clouds:
\begin{itemize}[nosep, leftmargin=*]
    \item {\em Statistical outlier removal (SOR) \cite{rusu2008towards}.} SOR can be used to remove local roughness on the (smooth) surface as a means of defense. SOR is not differentiable, producing obfuscated gradients for defenders. DUP-Net \cite{zhou2019dup} uses SOR and an upsampling network to reconstruct higher resolution point clouds. Similarly, IF-Defense \cite{wu2020if} utilizes SOR, followed by a geometry-aware model to encourage evenly distributed points. However, such defenders have been demonstrated to be attackable in \cite{tsai2020robust, ma2020efficient}. Dong \etal \cite{dong2020self} proposed replacing SOR by attention mechanism. 
    
    \item {\em Random sampling.} Yang \etal \cite{yang2019adversarial} suggested that 3D models with random sampling are robust to adversarial attacks. PointGuard \cite{liu2021pointguard} proposed majority voting for point cloud classification by predicting multiple randomly subsampled point clouds. 
    
    \item {\em Data augmentation.} Tramer \etal \cite{tramer2017ensemble} demonstrated that data augmentation can effectively account for adversarial attacks. Tu \etal \cite{tu2020physically} proposed generating physically realizable adversarial examples to train robust Lidar object detectors. Zhang \etal \cite{zhang2021art} proposed randomly permuting training data as a simple data augmentation strategy. PointCutMix \cite{zhang2021pointcutmix} pairs two training clouds and swaps some points between the pair to generate new training data.
\end{itemize}

\bfsection{Permutohedral lattice}
Permutohedral lattice is a powerful operation to project the coordinates from a high dimensional space onto a hyperplane that defines the lattice. It has been widely used in high dimensional filtering \cite{adams2010fast, jampani2016learning} that consists of three components, \ie splat, blur and slice. In particular, we illustrate the splat in Fig. \ref{fig:LPC}, where each square represents a projection (\ie projected point) from a 3D coordinate. The splat first locates the enclosing lattice simplex for the 3D point and calculates the vertex coordinates of the simplex. Then each projection distributes its value to the vertices using barycentric interpolation with barycentric weights that are calculated as the normalized triangular areas between the projection and any pair of its corresponding lattice vertices. 
We refer the readers to \cite{adams2010fast} for more details. Recently permutohedral lattice has been successfully explored in point cloud segmentation \cite{su2018splatnet, gu2019hplflownet, rosu2021latticenet} with remarkable performance.

\section{Robust Structured Declarative Classifiers}
\subsection{Structured Declarative Defender}
Recall that adversarial defenders often aim to denoise the input point clouds by reconstructing them in certain ways, and sparse coding \cite{zhang2015survey} is one of the classic approaches for finding a sparse representation of the input data in the form of a linear combination of basic elements as well as those basic elements themselves. Due to its simplicity, we consider using sparse coding as a means to construct declarative nodes.

Specifically, in the 3D space given a (learnable) dictionary $\mathbf{B}\in\mathbb{R}^{3\times N}$ with $N\gg 3$ atoms, (structured) sparse coding aims to solve the following optimization problem for each point $\mathbf{x}_i\in\mathbb{R}^3$ in a point cloud $\mathbf{x}=\{\mathbf{x}_i\}\subseteq\mathbb{R}^3$:
\begin{align}\label{eqn:sc}
    \Tilde{\mathbf{x}}_i \in \argmin_{\mathbf{z}\in\mathcal{Z}} f(\mathbf{x}_i, \mathbf{z}) = \frac{1}{2}\|\mathbf{x}_i - \mathbf{B}\mathbf{z}\|^2 + \phi(\mathbf{z}), 
\end{align}
where $\phi$ denotes a regularization term, and $\mathcal{Z}\subseteq\mathbb{R}^N$ denotes a structural feasible solution space.

\bfsection{Obfuscated \& implicit gradients in $\frac{\partial \Tilde{\mathbf{x}}_i}{\partial \mathbf{x}_i} = \frac{\partial \mathbf{z}}{\partial \mathbf{x}_i}\Big|_{\mathbf{z}=\Tilde{\mathbf{x}}_i}$}
To see this, we can rewrite Eq.~\eqref{eqn:sc} as $\Tilde{\mathbf{x}}_i \in \argmin_{\mathbf{z}} F(\mathbf{x}_i, \mathbf{z}) = f(\mathbf{x}_i, \mathbf{z}) + \phi(\mathbf{z}) + \delta(\mathbf{z})$, where $\delta(\mathbf{z})$ denotes the Dirac delta function returning 0 if $\mathbf{z}\in\mathcal{Z}$ holds, otherwise, $+\infty$. Therefore, $F$ will become non-differentiable when $\mathbf{z}\notin\mathcal{Z}$ or $\phi$ is non-differentiable over $\mathbf{z}\in\mathcal{Z}$, leading to obfuscated gradients. Otherwise, based on the first-order optimality condition, we have 
$\mathbf{B}^T\mathbf{B}\Tilde{\mathbf{x}}_i - \mathbf{B}^T\mathbf{x}_i + \phi'(\Tilde{\mathbf{x}}_i) = \mathbf{0}$, where $(\cdot)^T$ denotes the matrix transpose operator and $\phi'$ denotes the first-order derivative of $\phi$. By taking another derivative on both sides, we then have a linear system $\left(\mathbf{B}^T\mathbf{B}+\phi''(\Tilde{\mathbf{x}}_i)\right)\cdot\frac{\partial \Tilde{\mathbf{x}}_i}{\partial \mathbf{x}_i} = \mathbf{B}^T$, if the second-order derivative, $\phi''$, exists at $\mathbf{z}\in\mathcal{Z}$. Clearly, solving this linear system may be challenging, because $\phi''(\Tilde{\mathbf{x}}_i)$ may not be computable and the matrix $\mathbf{B}^T\mathbf{B}+\phi''(\Tilde{\mathbf{x}}_i)$ may be rank-deficient. Such phenomenons lead to implicit gradients.

\subsection{Parameter Learning via Bilevel Optimization} 
Fig. \ref{fig:framework} illustrates our approach with two components, \ie a declarative defender $f$ and a network backbone $g$ that takes the outputs of the defender as input and then makes predictions. Equivalently we can formulate the training of such networks as a bilevel optimization problem as follows:
\begin{align}
    & \min_{\mathbf{B}, \omega} \sum_{(\mathbf{x}, y)\in\mathcal{X}\times\mathcal{Y}} \ell\Big(g(\Tilde{\mathbf{x}}; \omega), y\Big), \\
    & \mbox{s.t.} \,
    \Tilde{\mathbf{x}}_i \in \argmin_{\mathbf{z}\in\mathcal{Z}}\frac{1}{2}\|\mathbf{x}_i - \mathbf{B}\mathbf{z}\|^2 + \phi(\mathbf{z}), \forall \mathbf{x}_i\in\mathbf{x}, \nonumber
\end{align}
where $(\mathbf{x}, y)\in\mathcal{X}\times\mathcal{Y}$ denotes a training sample with data $\mathbf{x}$ and label $y$, $\ell$ denotes a loss function such as cross-entropy, and $\mathbf{B}, \omega$ denote the defender and network parameters, respectively. Same as training deep networks, we can solve this optimization problem using (stochastic) gradient descent.


\bfsection{Validity of $\frac{\partial g}{\partial \Tilde{\mathbf{x}}_i} = \frac{\partial g}{\partial \mathbf{z}}\Big|_{\mathbf{z}=\Tilde{\mathbf{x}}_i}$ in the structural space}
In a gradient based adversarial attack, the gradient for modifying an input point $\mathbf{x}_i$ through backpropagation can be written as $\frac{\partial g}{\partial \Tilde{\mathbf{x}}_i}\frac{\partial \Tilde{\mathbf{x}}_i}{\partial \mathbf{x}_i}$. Assuming that $\frac{\partial \Tilde{\mathbf{x}}_i}{\partial \mathbf{x}_i}$ can be computed exactly, then the gradient in the attack would hold in general only if $\mathbf{z}$ was unconstrained so that $\frac{\partial g}{\partial \Tilde{\mathbf{x}}_i}$ is valid. Unfortunately in our case this is not true as we constrain $\mathbf{z}\in\mathcal{Z}$. Therefore, the attack in the declarative node will produce inaccurate gradients, and such errors will be propagated to the adversarial examples, leading to failure cases together with $\frac{\partial \Tilde{\mathbf{x}}_i}{\partial \mathbf{x}_i}$.


\subsection{Instantiation: Lattice Point Classifier}
So far we have explained the learning principles and defense philosophy in our approach. The key challenge now is how to design the structural space $\mathcal{Z}$ and the regularizer $\phi$ to achieve robust point cloud classifiers that can be trained and inferred effectively and efficiently. To address this issue, we borrow the idea from permutohedral lattice, and propose an instantiation, namely, Lattice Point Classifier (LPC). 

\setlength{\intextsep}{0pt}%
\setlength{\columnsep}{10pt}%
\begin{wrapfigure}{r}{.4\linewidth}
	\begin{minipage}[b]{\linewidth}
		\begin{center}
			\centerline{\includegraphics[width=\linewidth]{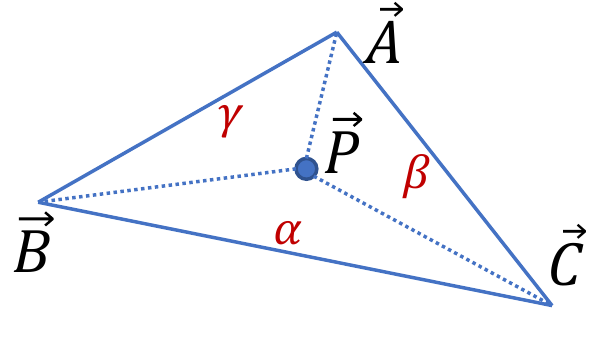}}
		\end{center}
	\end{minipage}
	\vspace{-11mm}
    \caption{Illustration of barycentric coordinates and weights.}
    \label{fig:bw}
\end{wrapfigure}

\bfsection{Geometric view on barycentric coordinates and their weights}
Barycentric coordinates ($\overrightarrow{A}$, $\overrightarrow{B}$, $\overrightarrow{C}$ in Fig. \ref{fig:bw}) can be used to express the position of any point ($\overrightarrow{P}$) located on the entire triangle with three scalar weights ($\alpha, \beta, \gamma$). To compute $\overrightarrow{P}$ using barycentric coordinates we can always use the following equation:
\begin{align}\label{eqn:P}
    \overrightarrow{P} = \alpha \overrightarrow{A} + \beta \overrightarrow{B} + \gamma \overrightarrow{C}, \, \exists \alpha, \beta, \gamma \geq 0, \alpha+\beta+\gamma = 1.
\end{align}
Note that this equation holds for an arbitrary dimensional space, including the 3D space. Now given $\overrightarrow{A}, \overrightarrow{B}, \overrightarrow{C}, \overrightarrow{P}$, we can compute $\alpha, \beta, \alpha$ using the normalized areas. Taking $\gamma$ for example, we can compute it as follows:
\begin{align}
    \gamma = \frac{\|\overrightarrow{AB}\times\overrightarrow{AP}\|}{\|\overrightarrow{AB}\times\overrightarrow{AC}\|} \propto \|\overrightarrow{AB}\times\overrightarrow{AP}\|,
\end{align}
where $\overrightarrow{AB}=\overrightarrow{B} - \overrightarrow{A}, \overrightarrow{AP}=\overrightarrow{P} - \overrightarrow{A}, \overrightarrow{AC}=\overrightarrow{C} - \overrightarrow{A}$, $\times$ denotes the cross product operator, and $\|\cdot\|$ denotes the $\ell_2$ norm of a vector measuring its length. Clearly, the barycentric weights define a nonlinear mapping that can be computed efficiently using the splat operation for the permutohedral lattice.

\bfsection{Constructing $\mathcal{Z}$ and $\phi$ using barycentric weights}
By substituting Eq. \eqref{eqn:P} into Eq. \eqref{eqn:sc}, we manage to define a structured sparse coding problem, where the structural space $\mathcal{Z}$ and the regularizer $\phi$ can be constructed as follows:
\begin{align}
    &\mathcal{Z}\stackrel{def}{=}\left\{\mathbf{z} \mid \mathbf{z}^T\mathbf{e} = 1, \mathbf{z} \succeq \mathbf{0} \right\}, \\ &\phi(\mathbf{z})\stackrel{def}{=} \lambda\sum_{n=1}^N \|\mathbf{B}\mathbf{z} - \mathbf{B}_n\|\cdot 1_{\{\mathbf{z}_n>0\}}, 
\end{align}
where $\mathbf{e}$ denotes a vector of 1's, $\succeq$ denotes the entry-wise operator of $\geq$, $\mathbf{B}_n\in\mathbb{R}^3$ denotes the $n$-th column in $\mathbf{B}$, $1_{\{\cdot\}}$ denotes a binary indicator returning 1 if the condition holds, otherwise 0 (\ie the {\em binarization} of barycentric weights), and $\lambda\geq0$ is a small constant controlling the contribution of $\phi$ to the objective so that it will not be dominated by $\phi$.

\begin{prop}\label{prop:1}
Supposing that $\mathbf{B}$ in Eq. \eqref{eqn:sc} represents the vertices in a permutohedral lattice that is large enough to cover all possible projections from points among the data, then there exists a solution to minimize the reconstruction loss using three vertices, at most, and the minimum loss is equal to the projection loss onto the lattice.
\end{prop}
This is because Eq. \eqref{eqn:P} defines a lossless representation using a linear combination of three vertices. The only loss occurs when projecting a point to the permutohedral lattice.

\setlength{\intextsep}{0pt}%
\setlength{\columnsep}{10pt}%
\begin{wrapfigure}{r}{.4\linewidth}
	\begin{minipage}[b]{\linewidth}
		\begin{center}
			\centerline{\includegraphics[width=\linewidth]{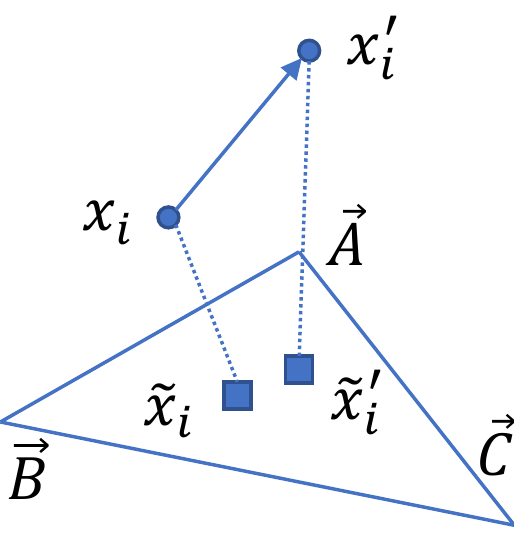}}
		\end{center}
	\end{minipage}
	\vspace{-7mm}
    \caption{Illustration of our defense mechanism.}
    \label{fig:defense}
\end{wrapfigure}
\bfsection{Defense mechanism in LPC}
Fig. \ref{fig:defense} illustrates how our declarative defender works with the permutohedral lattice, where the triangle, circles and squares represent a lattice cell, 3D points and their projections on the cell, respectively. In the adversarial point cloud, the attacker modifies a point from $\mathbf{x}_i$ to $\mathbf{x}_i'$. Then during the inference, our defender projects $\mathbf{x}_i'$ to $\Tilde{\mathbf{x}}_i'$ on the lattice.

\begin{prop}
Supposing $\Tilde{\mathbf{x}}_i = \alpha \overrightarrow{A} + \beta \overrightarrow{B} + \gamma \overrightarrow{C}$ where $\alpha, \beta, \gamma$ are the barycentric weights, then in order to guarantee that $\Tilde{\mathbf{x}}_i, \Tilde{\mathbf{x}}_i'$ lie in different cells, the distance between $\mathbf{x}_i$ and $\mathbf{x}_i'$ should be bigger than the shortest distance between $\Tilde{\mathbf{x}}_i'$ and the boundary of the triangle, that is:
\begin{align}
    \|\mathbf{x}_i - \mathbf{x}_i'\| > \min\left\{\frac{\alpha}{\|\overrightarrow{BC}\|}, \frac{\beta}{\|\overrightarrow{AC}\|}, \frac{\gamma}{\|\overrightarrow{AB}\|}\right\}\cdot s,
\end{align}
where $s$ denotes the area of the triangle. In particular, if the triangle is equilateral, then $\|\mathbf{x}_i - \mathbf{x}_i'\| > \frac{\sqrt3}{2}l\cdot\min\{\alpha, \beta, \gamma\}$ where $l$ denotes the side length.
\end{prop}
It will be more intuitive to understand this result if we binarize the barycentric weights before feeding them into the backbone network for classification, because different projections lying in the same lattice cell will lead to the same representation. This could be an effective way to remove the adversarial noise in the data. Also, this result indicates that (1) the points whose projections are closer to the boundary are easier to change their sparse representations, (2) the movements that make such changes are proportional to the scale of the lattice cell. In general, larger cells will be more tolerant to the adversarial noise, but they may sacrifice the generalization of the classifiers.

\begin{figure}[t]
    \centering
    \includegraphics[width=0.9\linewidth]{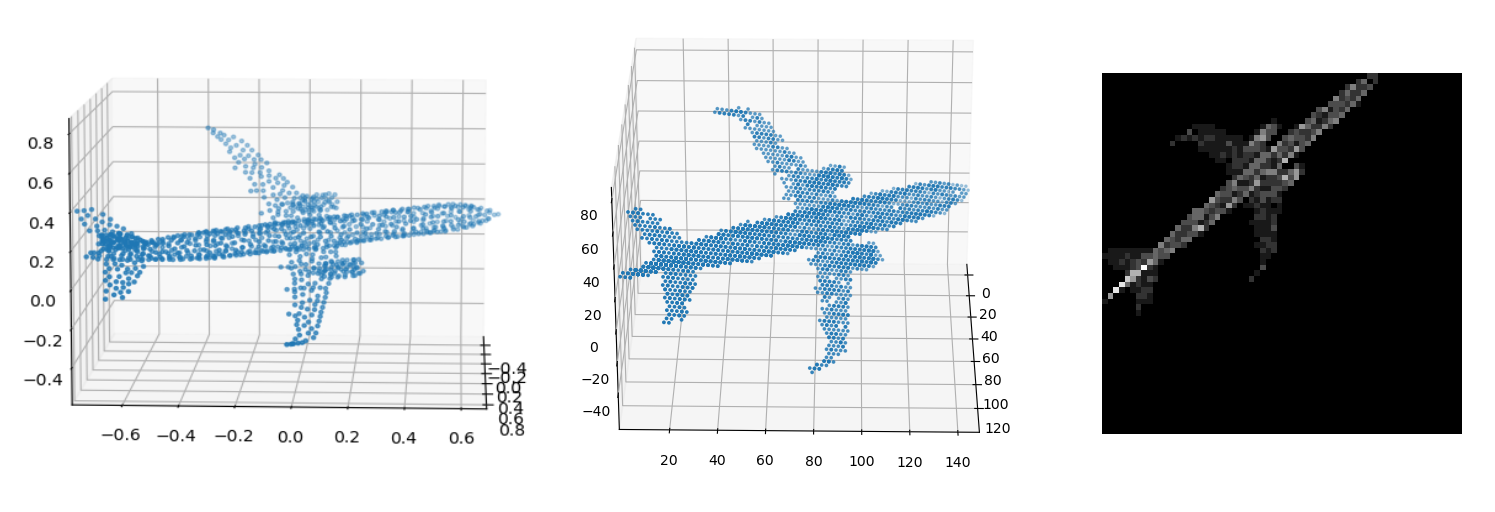}
    \vspace{-3mm}
    \caption{Illustration of {\bf (left)} a point cloud, {\bf (middle)} its lattice representation, {\bf (right)} its image for classification.}
    \label{fig:image}
  	\vspace{-4mm}
\end{figure}

\bfsection{Workflow of LPC}
To summarize, it is operated as follows:
\begin{enumerate}[nosep, leftmargin=*]
    \item Given a point cloud, the barycentric weights of each point are computed using the splat for permutohedral lattice;
    
    \item Generate an image for the point cloud by averaging the barycentric weights over all the points (after binarization if applied) and aligning the lattice with the image representation (see Fig. \ref{fig:image} for illustration);
    
    \item Apply a 2D CNN as the backbone network to classify the image produced by the point cloud.
\end{enumerate}

\bfsection{Implementation}
The key challenge in our implementation of LPC is how to determine the projection matrix for the hyperplane and the scale of each permutohedral lattice cell. 
\begin{itemize}[nosep, leftmargin=*]
    \item {\em Projection matrix:} By referring to \cite{adams2010fast}, we initialize the projection matrix as $\footnotesize \begin{bmatrix}
2  & -1 & -1\\
-1 &  2 & -1\\
-1 & -1 &  2\\
\end{bmatrix}$ and train the network in an end-to-end fashion. However, we observe that a big update of this matrix will make the training crash, and to avoid this issue, the update per iteration has to be tiny, leading to almost unnoticeable change eventually. The reason for this phenomenon is that this matrix has to satisfy certain requirements (see \cite{adams2010fast}), and thus the unconstrained update in backpropagation cannot work here. Therefore, in our experiments we initialize and fix the projection matrix. We refer to \cite{gu2019hplflownet} for the lattice transformation. 
    
    \item {\em Scale of lattice cell:} The parameter is simply not differentiable in backpropagation, and thus we tune it as a predefined hyper-parameter using cross-validation with grid search, same for the other hyper-parameters such as learning rate. Such scales have a significant impact on the image resolution used in 2D CNNs for classification.
\end{itemize}

Specifically, we evaluate our LPC comprehensively based on three different CNNs, \ie VGG16 \cite{simonyan2014very}, ResNet50 \cite{he2016deep}, and EfficientNet-B5 \cite{tan2019efficientnet} as the backbone network with randomly initialization. By default, the image resolution for each backbone network is $512\times512$, $128\times128$ and $456\times456$, respectively. On ModelNet40 \cite{wu20153d} we train the three models using a learning rate of $10^{-4}$, but on ScanNet \cite{dai2017scannet} we only train our best model, \ie EfficientNet-B5, using a learning rate of $5\times10^{-5}$. We use Adam \cite{kingma2014adam} as our optimizer in all of our experiments with weight decay of $10^{-4}$ and learning rate decay of 0.7 for every 20 epochs. Dropout \cite{srivastava2014dropout} and data augmentation are applied as well when needed.

\section{Experiments}



\bfsection{Datasets} 
We conduct our experiments on ModelNet40 \cite{wu20153d} and ScanNet \cite{dai2017scannet}. ModelNet40 has a collection of 12,311 3D CAD objects from 40 common categories. It is split into 9,843 training and 2,468 test samples. Following \cite{qi2017pointnet, xiang2019generating}, we uniformly sample 1,024 points from the surface of the original point cloud per object and scale them into a unit ball. ScanNet contains 1,513 RGB-D scans from over 707 real indoor scenes with 2.5 million views. Following \cite{li2018pointcnn}, we generate 12,445 training and 3,528 test point clouds from 17 categories, with 1,024 points for each point cloud as well.

\begin{figure}[t]
	\begin{minipage}[b]{0.495\linewidth}
		\begin{center}
			\centerline{\includegraphics[width=1.05\linewidth, keepaspectratio,]{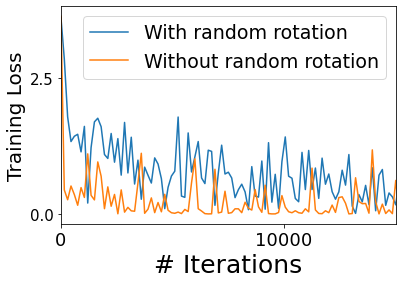}}
			\vspace{-1mm}
			\centerline{(a) ModelNet40}
		\end{center}
	\end{minipage}
	\hfill
	\begin{minipage}[b]{0.495\linewidth}
		\begin{center}
			\centerline{\includegraphics[width=1.01\linewidth,keepaspectratio]{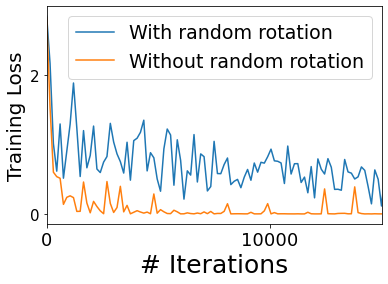}}
			\vspace{-1mm}
			\centerline{(b) ScanNet}
		\end{center}
	\end{minipage}

	\vspace{-5mm}
    \caption{Training loss comparison on both datasets using EfficientNet-B5 as the backbone.}
    \label{fig:benchmark}
	\vspace{-4mm}
\end{figure}

\bfsection{Baselines} 
We compare our Lattice Point Classifier (LPC) with different adversarial defenders for point clouds that work with PointNet \cite{qi2017pointnet}, including DUP-Net \cite{zhou2019dup} (with SOR and the upsampling network), IF-Defense \cite{wu2020if} (with ConvONet \cite{peng2020convolutional}), and robust pooling layer (RPL) \cite{gould2021deep}. We utilize public code \cite{xu2019pnetimp} to train PointNet using the default setting and evaluate DUP-Net and IF-Defense based on the implementation in \cite{wu2020if}. We report the best performance for each defender after fine-tuning. Specifically, we set $k=2$ (the number of neighbor points) in KNN and $\alpha=1.1$ (the percentage of outliers) for SOR, use 2 for the upsampling rate in DUP-Net, and choose the Welsch penalty function \cite{dennis1978techniques} for RPL \cite{gould2021deep} to replace the max pooling layer in PointNet. The model with RPL is trained with adversarial point clouds where 10\% of input points are replaced by random outliers.

\bfsection{Adversarial attackers} 
Eight attackers are utilized to evaluate the robustness of different point cloud classifiers, including untargeted attackers (FGSM \cite{goodfellow2014explaining} and JGBA \cite{ma2020efficient}), targeted attackers (perturbation, add, cluster and object attackers \cite{xiang2019generating}), isometry transformation based attackers \cite{zhao2020isometry} (the untargeted black-box TSI attack and the targeted white-box CTRI attack), and GAN based LG-GAN attacker \cite{zhou2020lg}. 

We apply both FGSM and JGBA to the full test set of both datasets. We slightly modify FGSM for attacking DUP-Net and IF-Defense under the white-box setting as both defenders are non-differentiable. To do so, during an attack we simulate the SOR process and obtain the indices of the remaining points based on which the gradient is passed to the remaining points. By default, the parameter $\epsilon$ in FGSM is set to 0.1, and the perturbation norm constraint $\epsilon$, number of iterations $n$ and step size $\alpha$ in JGBA are set to 0.1, 40, 0.01, respectively. Such parameters work well in practice.


For the targeted attackers, by following \cite{xiang2019generating} we pick 10 large classes from ModelNet40, where a batch of 6 point clouds per class is randomly selected and attacked using the other 9 classes as the targets, leading to $10 \times 9 \times 6 = 540$ victim-target pairs. Similarly, from ScanNet we randomly select a batch of 6 point clouds as well from the 7 classes that contain more than 100 point clouds in test data. The learning rate of all the targeted attackers is set to 0.01. For DUP-Net and IF-Defense, we first attack clean PointNet to obtain adversarial point clouds, and then feed them into DUP-Net and IF-Defense for prediction. The perturbation attack first introduces small random perturbations on all original points, and uses $L_2$ norm to constrain the adversarial shifts. 
Using the add attacker, we add 60 points to each cloud and use Chamfer distance as the metric. Cluster and object attackers generate the initial clusters with parameter $\epsilon=0.11$ in DBSCAN \cite{ester1996density}. 
Using the cluster attacker, we add 3 clusters of 32 points to the original clouds. Using the object attacker, three 64-point adversarial objects are attached to each original point cloud.

For the TSI/CTRI attacker in \cite{zhao2020isometry}, we evaluate the performance of LPC on ModelNet40 using EfficientNet-B5. By following \cite{zhao2020isometry} we use this attacker with the default settings to attack 2,000 randomly selected point clouds. The attacker applies the black-box TSI attacker first to trick the models, and then the white-box CTRI attacker if TSI fails.

We implemented a vanilla version of LPC in TensorFlow with binarized weights, ResNet50 as the backbone and $128\times128$ as the 2D image size. On ModelNet40, we trained a TensorFlow LPC model and integrated it into LG-GAN to generate adversarial samples. For the benchmark models, we trained a TensorFlow PointNet \cite{qi2017pointnet} model and generates adversarial samples with it since all other benchmark defenses are PointNet based. We followed the setups in \cite{zhou2020lg} and set weight factor $\alpha=100$.

\begin{table}
\setlength{\tabcolsep}{1pt}
\centering
\caption{Our learning choice comparison in terms of test accuracy (\%) with learning rate $10^{-4}$.}
\vspace{-3mm}
\begin{tabular}{  c | c  c  c  c  c  c }
    \toprule
    T-Net \cite{qi2017pointnet}           &      & \checkmark & & & \checkmark &            \\
    Binarized weights      &      &            & \checkmark &  & \checkmark & \checkmark \\
    Random rotation   &      &            &            &      \checkmark        &            & \checkmark   \\
    \midrule
    {\small ResNet-50 on ModelNet40}     & 88.2 &       87.3 &       \textbf{88.9} & 60.0 & 88.2  & 75.2 \\
    \midrule
    {\small EfficientNet-B5 on ModelNet40}     & - &       - &       \textbf{89.5} & 83.3 & - & 84.8 \\
    \midrule
    {\small EfficientNet-B5 on ScanNet}     & 80.5 &       - &      80.0 & {\bf 82.6} & - & - \\
\bottomrule
\end{tabular}
\label{table:cls_module}
\vspace{-2mm}
\end{table}

\begin{table}
\centering
\setlength{\tabcolsep}{4pt}
\caption{Inference Time (ms) on an NVIDIA V100S GPU with batch size 1 (clean PointNet \cite{qi2017pointnet} takes 8.6 ms)}
\vspace{-2mm}
\begin{tabular}{  c  c  c | c  c  c}
    \toprule
     \thead{DUP-Net\\\cite{zhou2019dup}} & \thead{IF-Defense\\\cite{wu2020if}} 
     & \thead{RPL\\\cite{gould2021deep}} & \thead{LPC w/\\VGG} & \thead{LPC w/\\ResNet} &  \thead{LPC w/\\EfficientNet} \\
    \midrule
	 808.1   &  1793.6 & 62.0 & 21.4 & 25.1 & 59.0 \\
    \bottomrule

\end{tabular}
\label{table:run_time}
\vspace{-3mm}
\end{table}

\begin{table*}[t]
\centering
\caption{Test accuracy (\%) comparison (higher is better) on the full test datasets, where ``-'' indicates no result.} \vspace{-3mm}
\begin{tabular}{  cc | c | ccc | ccc  }
    \toprule
         & & PointNet & \thead{PointNet w/\\DUP-Net \cite{zhou2019dup}} & \thead{PointNet w/\\IF-Defense \cite{wu2020if}} & \thead{PointNet w/\\ RPL \cite{gould2021deep}} & \thead{LPC w/\\ VGG16} & \thead{LPC w/\\ ResNet50} & \thead{LPC w/\\EfficientNet}  \\
    \hline
	\hline
	\multirow{3}{*}{ModelNet40} & No attack 
	& \({\bf 90.15}\) & \(89.30\) & \(87.60\) &  \(84.76\) & \(88.65\) & \(88.90\) & \(89.51\)  \\
    & FGSM \cite{goodfellow2014explaining}
	& \(45.99\) & \(61.63\) &  \(38.75\)&   \(0.04\) & \(88.65\) & \(88.90\) & \({\bf 89.51}\)   \\
    & JGBA \cite{ma2020efficient}
	&  \(0.00\) &  \(1.14\) &  \(5.37\) &   \(0.00\) & \(88.65\) & \(88.90\) & \({\bf 89.51}\)   \\
	\hline
	\hline
	\multirow{3}{*}{ScanNet} & No attack 
	& \({\bf 84.61}\) & \(83.62\) & \(80.19\) &  \(76.02\) & - & - & \(83.16\)   \\
    & FGSM \cite{goodfellow2014explaining}
	& \(45.66\) & \(73.67\) & \(71.14\) &   \(1.70\) &  - & - & \({\bf 83.16}\)    \\
    & JGBA \cite{ma2020efficient}
	&  \(0.00\) &  \(7.77\) & \(13.45\) &   \(0.00\) &  - & - & \({\bf 83.16}\)    \\
	\bottomrule

\end{tabular}
\label{table:attack_accu}
\end{table*}

\begin{table*}[t]
\centering
\caption{Attack success rate (\%) comparison (lower is better), where ``-'' indicates no result. The standard deviations of our LPC range from 0\% to 0.28\% in success rate on ModelNet40.}\vspace{-3mm}
\begin{tabular}{ c c | c | c  c  c | c  c  c  }
    \toprule
         & & PointNet & \thead{PointNet w/\\DUP-Net \cite{zhou2019dup}} & \thead{PointNet w/\\IF-Defense \cite{wu2020if}} & \thead{PointNet w/\\ RPL \cite{gould2021deep}} & \thead{LPC w/\\ VGG16} & \thead{LPC w/\\ ResNet50} & \thead{LPC w/\\EfficientNet}  \\
    \hline
    \hline
    \multirow{6}{*}{ModelNet40} & FGSM \cite{goodfellow2014explaining}
	& \(48.99\) & \(30.77\) &  \(55.78\)&  \(99.95\)&  \({\bf 0.00}\) &  \({\bf 0.00}\) &  \({\bf 0.00}\)   \\
    & JGBA \cite{ma2020efficient}
	& \(100.00\)& \(98.73\) &  \(93.85\) & \(100.00\)&  \({\bf 0.00}\) &  \({\bf 0.00}\) &  \({\bf 0.00}\)   \\
	& Perturbation \cite{xiang2019generating}
	& \(100.00\)&  \({\bf 0.095}\) &  \({\bf 0.095}\) &  \(3.95\) &  \(0.56\) &  \(0.37\) &  \(0.38\)   \\
	& Add \cite{xiang2019generating}
	& \(99.72\) &  \({\bf 0.095}\) &  \({\bf 0.095}\) &  \(3.33\) &  \(0.56\) &  \(0.19\) &  \(0.19\)   \\
	& Cluster \cite{xiang2019generating}
	& \(98.34\) &  \(6.76\) &  \(6.30 \) & \(17.04\) &  \(1.11\) &  \({\bf 0.93}\) &  \(1.21 \)   \\
	& Object \cite{xiang2019generating}
	& \(98.43 \) &  \(1.02\) &  \(1.11\) & \(74.07\) &  \(0.93\) &  \(1.11\) &  \({\bf 0.75} \)   \\
	\hline
	\hline
	\multirow{6}{*}{ScanNet} & FGSM \cite{goodfellow2014explaining}
	& \(46.03\) & \(10.29\) & \(11.28\) &  \(97.76\)& - & - & \({\bf 0.00}\)    \\
    & JGBA \cite{ma2020efficient}
	& \(100.00\)& \(90.55\) & \(83.21\) & \(100.00\)& - & - & \({\bf 0.00}\)    \\
	& Perturbation \cite{xiang2019generating}
	& \(100.00\)&  \(3.17\) &  \({\bf 2.38}\) &  \(3.03\) & - & - & \(12.70\)    \\
	& Add \cite{xiang2019generating}
	& \(100.00\)&  \({\bf 1.98}\) &  \(2.38\) & \(18.65\) & - & - & \(2.78\)    \\
	& Cluster \cite{xiang2019generating}
	& \(100.00\)&  \(30.16\)&  \(23.81\)&  \(40.87\) & - & - & \({\bf 9.92}\)    \\
	& Object \cite{xiang2019generating}
	& \(100.00\)&  \(7.14\) &  \(7.54\) &  \(85.71\) & - & - & \({\bf 5.95}\)    \\
    \bottomrule

\end{tabular}
\label{table:attack_main}
\vspace{-4mm}
\end{table*}

\begin{table}
\centering
\setlength{\tabcolsep}{5pt}
\caption{Performance comparisons under LG-GAN \cite{zhou2020lg} attacks. Attack success rate (Succ., \%) on different models and their classification accuracy (Accu., \%) under attack}
\vspace{-2mm}
\begin{tabular}{ c | c | c  c  c | c }
    \toprule
      & \thead{PointNet\\\cite{qi2017pointnet}} & \thead{DUP-Net\\\cite{zhou2019dup}} & \thead{IF-Defense\\\cite{wu2020if}} 
     & \thead{RPL\\\cite{gould2021deep}} & \thead{LPC w/\\ResNet} \\
    \midrule
	 Accu. & 0.6  &  31.6 & 37.2 & 56.8 & {\bf76.7}  \\
	 Succ. & 97.0 &  13.9 & 10.4 &  2.6 &  {\bf0.8}  \\
    \bottomrule

\end{tabular}
\vspace{-1mm}
\label{table:lggan_attack}
\end{table}

\subsection{Ablation Study}
\bfsection{Learning choices}
In Table \ref{table:cls_module} we list three learning choices that we would like to evaluate for improving the performance of vanilla LPC, \ie T-Net used in PointNet, binarized barycentric weights, and random rotation in data augmentation (together with point cloud random shifting and dropping \cite{qi2017pointnet}). We can see that: (1) T-Net seems to deteriorate the performance always. (2) Binarized weights work better on ModelNet40 than ScanNet, but compared with using barycentric weights the difference is <1\%. (3) Random rotation improves the performance on ScanNet, but worsens it on ModelNet40. This phenomenon can be partially explained from the training loss curves as shown in Fig. \ref{fig:benchmark}, where on ScanNet the overfitting occurs clearly without random rotation while on ModelNet40 random rotation makes the convergence slower and unstable. 

\bfsection{Running time}
Overall LPC achieves the fastest inference speed among all competing defenses as shown in Table \ref{table:run_time}. Our deepest model (17fps) is over 13 times faster than SOR based defenses. We can also improve the running time using shallower networks such as VGG16 (45fps) by slightly sacrificing the performance. We also tested the running time for each key component (declarative node and backbone 2D networks). With EfficientNet-B5 as the backbone, the declarative node and 2D network take $17.1$ and $41.9$ ms respectively during inference, and $15.5$ and $165.3$ ms during backpropagation. Clearly, the gradient passing through the declarative node takes relatively constant time in both feedforward and backpropagation. Considering the depth of EfficientNet-B5, the time spent on the declarative node is actually pretty long (recall that there is no learnable parameter inside), especially during inference. This partially validates our intuition of defending the adversarial attacks using implicit gradients.


\subsection{State-of-the-art Performance Comparison} \label{sect:cls_res}
We measure the robustness of classifiers using two metrics, \ie classification accuracy, and attack success rate. Classification accuracy under attacks is measured by feeding the entire adversarial attacked test set to the victim models. It is only calculated for untargeted attackers. The attack success rate is the ratio between the number of successful attacks to the number of all attempt attacks. For untargeted attacks, tricking the model to predict a wrong class is considered a success, while for targeted attacks it will have to trick the victim model to a specific class to be considered as successful. Untargeted attackers will only attack the point clouds that are correctly classified by the victim models, and targeted attackers will attack victim-target pairs.

\subsubsection{Classification Accuracy}
We summarize our comparison in Table \ref{table:attack_accu}. On ScanNet, we only show the performance of LPC with EfficientNet-B5, because it achieves the best accuracy on ModelNet40. We can see that: (1) On the clean test data with no attack, PointNet outperforms all the robust classifiers by small margins. (2) Using both attackers, our LPC variants work consistently and significantly better than the other defender-based robust classifiers as well as vanilla PointNet by large margins. For instance, compared with PointNet with IF-Defense under the JGBA attacks, the performance gaps are 84.14\% and 69.71\% on ModelNet40 and ScanNet, respectively. (3) For the backbone networks in LPC, it seems that the difference in performance is small among different CNNs. Though more evaluations are needed to confirm this, it also demonstrates the robustness of our approach.

\begin{figure}
    \center
    \includegraphics[width=.9\columnwidth]{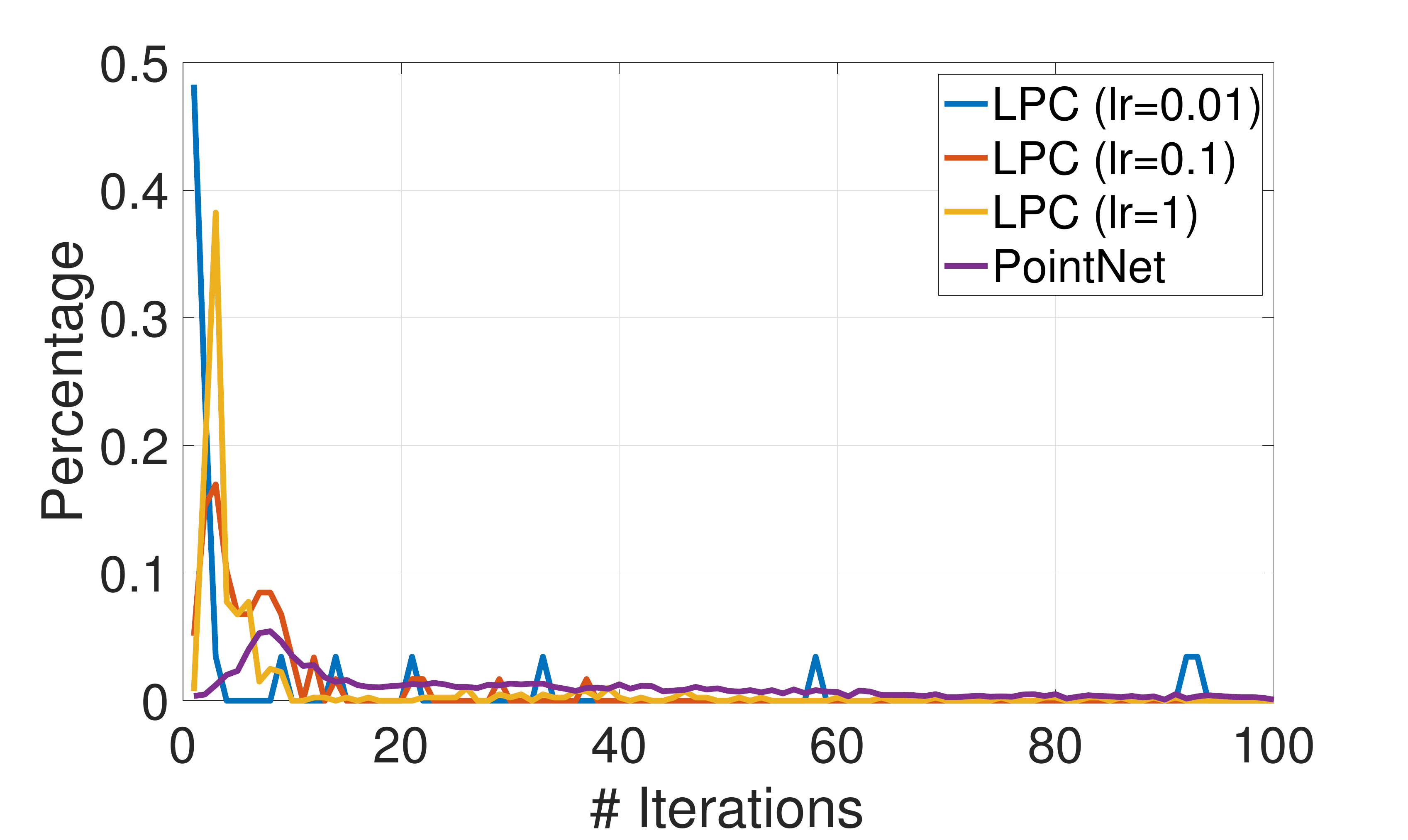}
    \vspace{-3mm}
	\caption{Distribution comparison of successful perturbation attacks on ModelNet40. To avoid the sparsity, LPC statistics are collected based on the cases from all the three models.}
	\label{fig:iter_num}
	\vspace{-4mm}
\end{figure}

\subsubsection{Attack Success Rate}

\bfsection{Performance summary}
We list our comparison in Table \ref{table:attack_main} and Table \ref{table:lggan_attack}. We can see that: (1) Our LPC variants achieve perfect results under both FGSM and JGBA attacks. Notice that compared with the other attackers, these two attackers aim to find adversarial examples fully based on gradients with no sampling. Their failure again strongly demonstrates the power of implicit gradients in defending gradient based adversarial attacks. (2) The SOR defense mechanism seems to work better for the perturbation and add attackers, but worse for the cluster and object attackers, compared with our LPC. However, overall LPC still achieves the best performance. 
(3) Our LPC also outperforms other defenses under GAN based attacks. Interestingly RPL \cite{gould2021deep} outperforms SOR defenses under LG-GAN attack, while under gradient based attacks the SOR methods are better. It implies LG-GAN has better transferability. The two SOR based defenses are gray-box attacks (partial access to the model). Gradient based adversarial samples suffer significant success rate drop as shown in Table \ref{table:attack_main}. But LG-GAN transfers well to SOR based defenses, maintaining over $10\%$ attack success rate.

Using the TSI/CTRI attacker \cite{zhao2020isometry}, our LPC with EfficientNet-B5 achieves the same success rate for both TSI and CTRI attackers on ModelNet40. Without random rotation, the result is 99.54\%, but with random rotation, the performance decreases to 67.33\% which is significantly better than PointNet (99.50\% and 99.55\% for TSI and CTRI, respectively). Such results also demonstrate that random rotation to point clouds as a means of data augmentation improves not only accuracy but also model robustness.

\begin{figure}
    \center
    \includegraphics[width=\columnwidth]{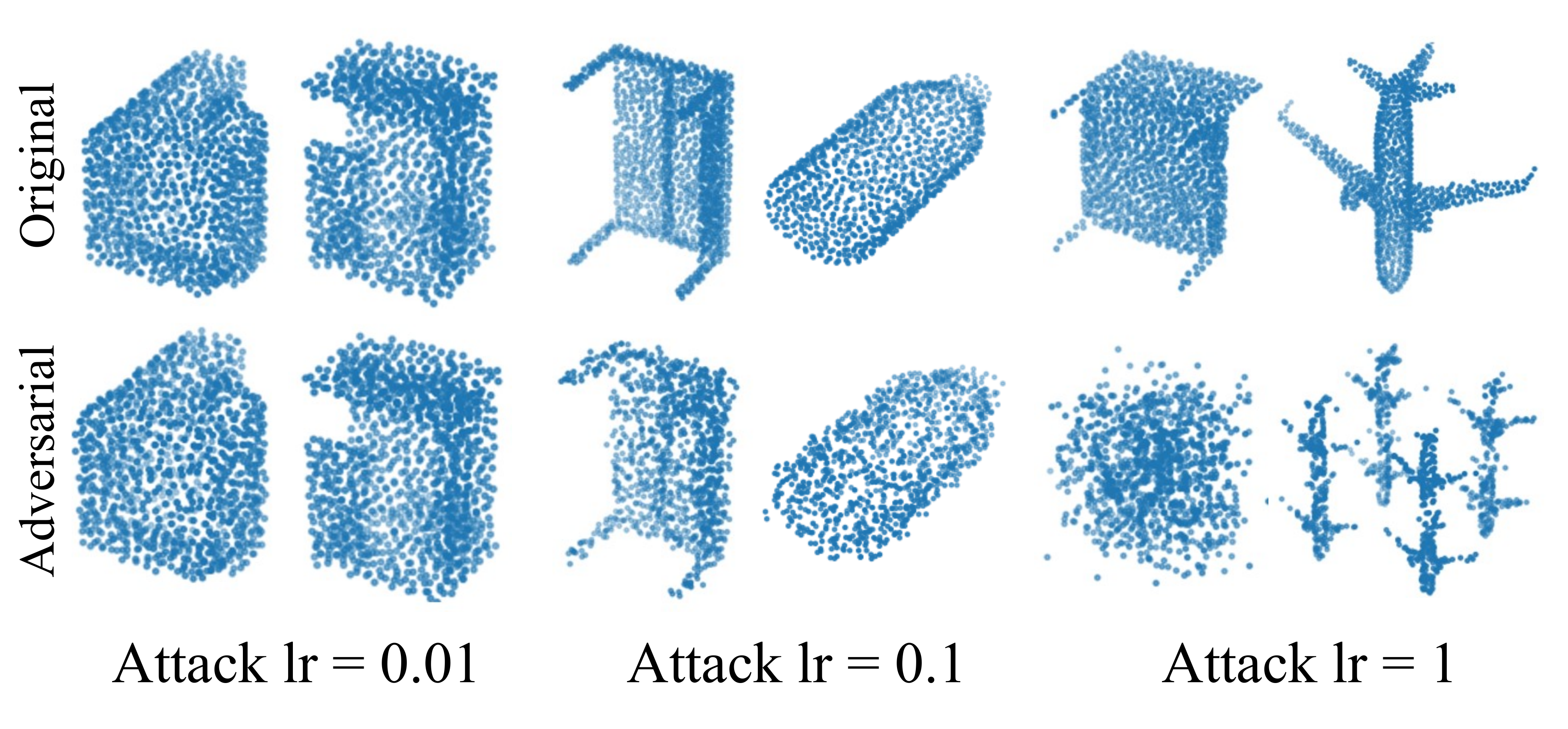}
    \vspace{-8mm}
	\caption{Successful adversarial perturbation examples on ModelNet40 for LPC with EfficientNet-B5.}
	\label{fig:succ_sample}
	\vspace{-4mm}
\end{figure}

\bfsection{Importance of gradients in finding successful adversarial examples for LPC}
To attack each point cloud, the four targeted attackers in \cite{xiang2019generating} conduct 10 random searches where 500 iterations are done to find the optimal adversarial samples. We notice that for PointNet, the best attacks could occur at any time within these 500 iterations, but for our LPC most optimal attacks happen at the first few iterations, as shown in Fig.~\ref{fig:iter_num} with different learning rates for the perturbation attacker. This behavior indicates that, in order to attack LPC, random search (a sampling based method) has contributed more to most of the successful attacks, rather than gradient-based iterations. With the increase of the learning rate, the gradients start to find more optimal adversarial samples. However, as we illustrate in Fig. \ref{fig:succ_sample}, with larger learning rates, the adversarial samples will not look similar to the original point clouds. For instance, with learning rate of 1, the point cloud of airplane has been totally changed to a mixture of smaller airplane point clouds, which is not an adversarial attack anymore. To sum up, such analysis again demonstrates the great potential of implicit gradients in defending adversarial attacks.



\section{Conclusion}
In this paper, we aim to address the problem of robust 3D point cloud classification by proposing a family of novel robust structured declarative classifiers, where a declarative node is defined by a constrained optimization problem such as the reconstruction of point clouds. The key insight in our approach is that the implicit gradients through the declarative node can help defend the adversarial attacks by leading them to wrong updating directions for inputs. We formulate the learning of our classifiers based on bilevel optimization, and further propose an effective and efficient instantiation, namely, Lattice Point Classifier (LPC). The declarative node in LPC is defined as structured sparse coding in the permutohedral lattice, whose outputs, \ie barycentric weights, are further transformed into images for classification using 2D CNNs. LPC is end-to-end trainable, and achieves state-of-the-art performance on robust classification on ModelNet40 and ScanNet using seven different adversarial attackers.

\bfsection{Limitations}
Currently the projection in the permutohedral lattice transformation in LPC is not learned but simply fixed as initialization. Also more evaluations for demonstrating the robustness of our approach across different backbone networks and datasets are desirable. Therefore, in our future work we will investigate more on how to properly learn the projection with its physical conditions and conduct more experiments to further demonstrate our robustness.

\section*{Acknowledgement}
K. Li and C. Zhong were supported in part by USDA NIFA under the award no. 2019-67021-28996. Z. Zhang was supported in part by NSF grant CCF-2006738. G. Wang was partly supported by NSERC grant RGPIN-2021-04244.

\balance
{\small
\bibliographystyle{ieee_fullname}
\bibliography{egbib}
}

\end{document}